\documentclass[journal]{IEEEtran}
\usepackage{hyperref}       
\usepackage{url}            
\usepackage{booktabs}       
\usepackage{amsfonts}       
\usepackage{nicefrac}       
\usepackage{microtype}      

\usepackage{amsmath}
\allowdisplaybreaks[4]
\usepackage{amssymb}
\usepackage{graphicx}
\usepackage{subfigure}
\usepackage{graphics}
\usepackage{multirow}
\usepackage{float}
\usepackage{bm}
\usepackage{mathrsfs}
\usepackage{color,soul}
\usepackage{algorithm}
\usepackage{algpseudocode}
\usepackage{color,soul}
\usepackage{enumerate}
\usepackage{todonotes}
\usepackage{latexsym}
\usepackage{wrapfig} 
\usepackage{ulem}
\usepackage{dsfont}
\sethlcolor{yellow}

\ifCLASSINFOpdf
\else
\fi

\begin{document}
%
\title{Nonlinear Collaborative Scheme for Deep Neural Networks}
%
%
%

\author{Hui-Ling Zhen, Xi Lin, Alan Z. Tang, Zhenhua Li,
        Qingfu Zhang,~\IEEEmembership{IEEE~Fellow} \\
        and Sam Kwong,~\IEEEmembership{IEEE~Fellow}  
\thanks{Hui-Ling Zhen, Xi Lin, Qingfu Zhang and Sam Kwong are with the Department
of Computer Science, City University of Hong Kong.}
\thanks{Zhenhua Li is with National University of Singapore, Institute of Operation Research. }
\thanks{Manuscript received April 19, 2005; revised August 26, 2015.}}

\maketitle

\begin{abstract}
Conventional research attributes the improvements of generalization ability of deep neural networks either to powerful optimizers or the new network design. Different from them, in this paper, we aim to link the generalization ability of a deep network to optimizing a new objective function. To this end, we propose a \textit{nonlinear collaborative scheme} for deep network training, with the key technique as combining different loss functions in a nonlinear manner. We find that after adaptively tuning the weights of different loss functions, the proposed objective function can efficiently guide the optimization process. What is more, we demonstrate that, from the mathematical perspective, the nonlinear collaborative scheme can lead to (i) smaller KL divergence with respect to optimal solutions; (ii) data-driven stochastic gradient descent; (iii) tighter PAC-Bayes bound. We also prove that its advantage can be strengthened by nonlinearity increasing. To some extent, we bridge the gap between learning (i.e., minimizing the new objective function) and generalization (i.e., minimizing a PAC-Bayes bound) in the new scheme. We also interpret our findings through the experiments on Residual Networks and DenseNet, showing that our new scheme performs superior to single-loss and multi-loss schemes no matter with randomization or not.  
\end{abstract}


%
\IEEEpeerreviewmaketitle

\section{Introduction}\label{sec:intro}

\IEEEPARstart{D}{eep} neural networks (DNNs), several years since their launch, have persistently enabled significant improvements in many application domains, such as pedestrian detection~\cite{sun2017face,zhang2016faster}, speech recognition~\cite{xiong2017microsoft} and natural language processing~\cite{kumar2016ask}, to name a few. Throughout these research, although learning is regarded as a central topic, generalization ability on the unseen test set is an ultimate goal of more significance~\cite{lecun2015deep}. However, in practice, the generalization ability is hard to measure during the optimization process, and it is commonly believed that there exists a huge gap between them. 

Thu, finding ways to decrease the gap between generalization and optimization is of both theoretical and practical importance. Throughout the given literature, there are two main research directions within this domain:

\begin{itemize}
\item Most of the works focus on investigating powerful optimizers on training set then checking the generalization ability on test set. Several algorithms have been proposed in this direction, such as Batch Normalization (BN)~\cite{bn2015batch}, dropout~\cite{dropout2014d}, to name a few. 

\item Some other research aims at changing network structures to improve the generalization ability, ignoring the optimization results. For example, residual networks~\cite{he2016deep} and highway networks~\cite{srivastava2015highway} have been constructed. 
\end{itemize}
To the best of our knowledge, little work has been done on the objective function, or called surrogate on the training set, except for the classical regularization term~\cite{lecun2015deep}. Different from the main research directions, in this paper, we propose a nonlinear collaborative training scheme based on \textit{new surrogate}, rather than designing a new optimizer/network structure.


Considering mentioned above, this paper proposes a \textit{nonlinear collaborative scheme} for network training, with a new surrogate for $0-1$ loss function and mathematical demonstration for generalization ability given. In the rest of this paper, we first describe the background and relative research for network training and generalization ability in Section \ref{sec:background}. Then our new scheme and demonstrations make up Sections \ref{sec:surrogate}-~\ref{sec:fourier}. Specifically,  

\begin{itemize}
	\item We construct a nonlinear collaborative objective function as a better surrogate loss in Section \ref{sec:surrogate}. The advantage of the new algorithmic scheme from the view of optimization is demonstrated in Section \ref{sec:alg}, with smaller KL divergence, data-driven stochastic gradient descent and faster convergence.
	\item We prove that the nonlinear collaboration scheme leads to better generalization ability in Section \ref{sec:theo}, by investigating a tighter PAC-Bayes bound. 
	\item With the help of spectral analysis, we demonstrate that the advantage of our new scheme can be strengthened with $p$ increasing in Section \ref{sec:fourier}. 
\end{itemize}
Further, our experiments, with different levels of randomization, on Residual Networks and Densenet are shown in Section \ref{sec:exp}, showing the better performance on train and validation accuracy on CIFAR-10 and CIFAR-100. Finally, our conclusions are drawn in Section \ref{sec:con}. 






\section{Background and Relative Work}\label{sec:background}

\subsection{Background}
For instance, in a classification problem, the generalization ability can be formulated as the $0-1$ loss function, 
\begin{align}\label{eq:zeroone}
\mathcal{L}_{0-1}(\textbf{w}) =  \frac{1}{n} \sum_{i=1}^{n} \mathds{1} [y^{(i)} \neq f(x^{(i)}|\textbf{w})],  
\end{align}
which is theoretically NP-hard in the training procedure~\cite{blum1989training,bartlett2006convexity,shalev2014understanding}. In practice, we always minimize predefined differentiable loss function based on training data~\cite{lecun2015deep}, which works as a surrogate from optimization perspective, rather than $0-1$ loss function directly\cite{lecun2015deep,hertz2018introduction,zhang2016understanding}. Some popular surrogate losses such as the mean square error (MSE),
\begin{align}\label{eq:mse}
\mathcal{L}_{MSE}(\textbf{w}) =  \frac{1}{n} \sum_{i=1}^{n} [y^{(i)},f(x^{(i)}|\textbf{w})]^2,
\end{align}
and the cross entropy loss (CE),
\begin{align}\label{eq:ce}
\mathcal{L}_{CE}(\textbf{w}) &=  -\frac{1}{n}  \sum_{i=1}^{n} [y^{(i)}\log f(x^{(i)}|\textbf{w})  \nonumber \\
&+ (1-y_i) \log (1- f(x^{(i)}|\textbf{w}))].
\end{align}
Therefore, \textbf{the gap firstly occurs in the relaxation between surrogate and $0-1$ loss function}.

For another, in network training, the test accuracy, or called a \textit{minimal risk}, 
 \begin{align}\label{eq:risk}
 R_{\mathcal{D}}(Q) = \mathbb{E}_{\mathbf{w} \sim Q}[R_{\mathcal{D}(\mathbf{w})}] = \mathbb{E}_{z \sim \mathcal{D}}[\mathbb{E}_{\mathbf{w} \sim Q}[\ell(\mathbf{w},z)]],
 \end{align}
is unknown during training. It is usually replaced by a train accuracy, or called \textit{empirical risk} 
\begin{align}\label{eq:emp_risk}
\widehat{R}_S(Q) & = R_{\widehat{\mathcal{D}}}(Q) = \mathbb{E}_{z \sim \widehat{ \mathcal{D}}}[\mathbb{E}_{\mathbf{w} \sim Q}[\ell(\mathbf{w},z)]]  \nonumber \\
& = \int \mathbb{E}_{\mathbf{w} \sim Q}[\ell(\mathbf{w},z_i)] \mathrm{d} P_{\widehat{ \mathcal{D}}}(\mathbf{w},z) \nonumber \\
& = \frac{1}{m} \sum_{i=1}^m \mathbb{E}_{\mathbf{w} \sim Q}[\ell(\mathbf{w},z_i)],
\end{align}
where $Z$ is a measurable space, $\mathcal{D}$ is an unknown distribution on $Z$, $\ell(\mathbf{w},z)$ denotes the loss function in the batch supervised learning, $S =\{z_1,z_2,\cdots,z_k \} \sim \mathcal{D}^k$ denotes the observation, $Q$ represents the weight distribution, $\widehat{\mathcal{D}}=\frac{1}{k} \sum\limits_{i=1}^k \delta_{z_i}$ is empirical distribution, respectively. Thus the \textit{generalization error} can be expressed as $G = R_{\mathcal{D}}(Q)-\widehat{R}_S(Q)$ when $Q$ is random and dependent on data $S$. Thus, it can be seen that \textbf{the gap also occurs in the difference between training and test accuracy}~\cite{hertz2018introduction,catoni2007pac,langford2002quantitatively}.

\subsection{Relative Work}
\subsubsection{Optimization}

Two main research lines lie within the recent literatures. (i) One is to propose \textit{powerful optimizers}. Except for BN and dropout given above, some accelerated techniques have also been investigated. For example, to accelerate the stochastic gradient descent (SGD), people have proposed stochastic variance reduction gradient (SVRG)~\cite{johnson2013accelerating} and its several variants~\cite{allen2016improved,kingma2014adam}, especially for the strongly non-convex assumption~\cite{allen2017natasha}. (ii) The other one is to construct novel network architectures. Apart from residual/highway networks mentioned above, some other network designs, such as dense networks~\cite{iandola2014densenet} for object detect/classification and $3$-D U-Net~\cite{u_net20163d} for volumetric segmentation from sparse annotation, have been investigated for specific tasks in deep or machine learning domain. We label all of these research as \textit{single-loss scheme} in this paper. Apparently, the objective functions used in single-loss scheme are irrelevant to the choice of the loss functions~\cite{steinwart2007compare}.

Considering that different loss functions have different theoretical properties \cite{golik2013cross,janocha2017loss} as well as different experimental performance \cite{zhao2017loss}, \textit{``multi-loss scheme"} has been utilized to break the limitations of single-loss scheme. The key technique is to linearly combined multiple loss functions via a fine-tuned approach for more robust performance~\cite{xu2016multi}. Experiments show the priority of multi-loss scheme on such complex environments as person re-identification~\cite{li2017person} and multi-label learning~\cite{sohn2016improved}.

Although the two schemes aim at investigating an appropriate approximation for the $0-1$ loss function, their drawbacks are apparent: in single-loss scheme, the objective function is irrelevant to the choice of loss function, while the effectiveness of multi-loss scheme is limited to the disadvantages of linear combination \cite{nonlinear2016regularized,nonlinear2017nonlinear}.

\subsubsection{Generalization}

Conventional research on generalization has centered on Probably Approximately Correct (PAC) learning \cite{denis1998pac,shalev2013stochastic} and PAC-Bayes learning \cite{catoni2007pac,langford2002quantitatively,langford2003pac}. Among them, PAC-Bayes bounds are a generalization of the Occam’s razor bound for algorithms which output a distribution over classifiers rather than just a single classifier \cite{catoni2007pac,langford2002quantitatively}. Moreover, PAC-Bayes bounds are much tighter (in practice) than most common VC-dimension related approaches on continuous classifier spaces \cite{langford2003pac}. Utilizing differentially private technique, \cite{sgd_pac2017entropy} has linked minimizing a local entropy \cite{entropy_sgd} to optimizing a PAC-Bayes data-dependent prior.

Limitations in the research mentioned above, clearly, are derived from the ignorance of relationship between optimization and generalization. To our best knowledge, the recent development mainly focuses on ``entropy-SGD", whose intended goal is for robust network training~\cite{entropy_sgd} while it has been found that the relationship between optimizing the entropy of network and minimizing PAC-Bayes bound~\cite{sgd_pac2017entropy}. Frankly, it is not the first attempt to improve generalization ability through optimization. In practice, we always add a regularization term during network training, which modify the loss function essentially. 


\subsection{Our Improvements}

Considering above, in this paper, we propose a \textit{nonlinear collaborative scheme}, aiming to investigate \textit{ a better surrogate} for $0-1$ loss function and improve the generalization ability during minimizing new objective function. Obviously, our key technique lies within the following two points:

\begin{itemize}
    \item We combine different loss functions, like MSE, CE, etc., in a nonlinear manner to construct a new surrogate on training set, unlike the separate utilization in single-loss scheme and linear combination in multi-loss scheme.    
    
    \item We demonstrate its advantage from both optimization and generalization perspectives, finding the new connection between optimizing new surrogate and minimizing PAC-Bayes bound induced by nonlinearity. 
\end{itemize}
Note that our work is different from \cite{sgd_pac2017entropy}, since its emphasis has been stressed on the modification of optimizer, rather than objective function. 



\section{Nonlinear Collaborative Scheme}\label{sec:surrogate}

\subsection{Algorithmic Scheme}

In this paper, our main contributions are to propose a \textit{nonlinear collaborative scheme} for network training, in which we utilize different loss functions in a nonlinear manner and generalization ability is geared towards optimization.

After viewing the learning problem as an optimization one, we mainly propose to \textit{construct a new surrogate on training set for $0-1$ loss function as the new objective function}
\begin{align}\label{eq:non_obj}
\mathcal{L}(x) = \left[ \sum_{m=1}^N \beta_m \mathcal{L}_m(x)^p \right]^{\frac{1}{p}},  
\end{align}
where each $\mathcal{L}_m(x)$ can be chosen from MSE, CE, Jensen–Shannon Divergence (JSD), etc., the power $p$ is fixed in the beginning of optimization, and $\beta_m$ ($m=1,2,\cdots,N$) parameterizes the nonlinear combination. In the following, we can also give a constraint for all $\beta_m$: $\beta_1+\beta_2+\cdots+\beta_N=1$. Based on~\ref{eq:non_obj}, we investigate the pseudocode in Algorithm \ref{alg:non_collaboration}, exhibiting the main procedures in our new scheme.

There are some tips in our proposed scheme: (i) Our key point is not the improvement of optimizer, thus it can be chosen from such given ones as Stochastic gradient descent, Adam \cite{kingma2014adam} and Adagrad \cite{duchi2011adaptive}. (ii) Different from multi-loss scheme, our another emphasis is the nonlinearity in new surrogate. In the following, we will demonstrate that utilizing the new surrogate as the objective function yields a better generalization on unseen test set.


\begin{algorithm}[H]%
			\caption{Nonlinear Collaborative Scheme}
			\label{alg:non_collaboration}%
			\begin{algorithmic}[1]%
				\item[]%
				\State Choose different loss functions to construct new nonlinear collaborative objective function and power $p$. 
				\State Initialization all the hyperparameters, including $\beta_{j}$ ($j=1,\cdots,N$), iteration $m$, epoch $s$, noise level $\epsilon$. Then choose an optimizer. 
				\ForAll {$1 \leq i \leq N$}
				\State Optimize the objective function
				\State Adaptive $\beta_j$ 	
				\EndFor
			\end{algorithmic}
\end{algorithm}

Hereby, the \textit{adaptive tuning rule} can be given as (Take $N=2$ as an example):
\begin{align}\label{eq:adaptive_weight}
& \beta_1= \max \left\{\frac{\exp(-\| \nabla \mathcal{L}_1(x) \| )}{Z}, \frac{\exp(-\| \nabla \mathcal{L}_2(x)\|)}{Z}\right\}, \nonumber \\
& \beta_2=1-\beta_1,  \nonumber
\end{align}
with $Z  =\exp(-\Vert \nabla \mathcal{L}_1(x) \Vert)+\exp(-\Vert \nabla \mathcal{L}_2(x) \Vert)$ as the normalization. It means that in every optimization step, we always choose the better gradient between $\mathcal{L}_1(x)$ and $\mathcal{L}_2(x)$. 

\subsection{Intuitive Interpretation}

\begin{figure}[H]
		\includegraphics[width=0.9 \linewidth]{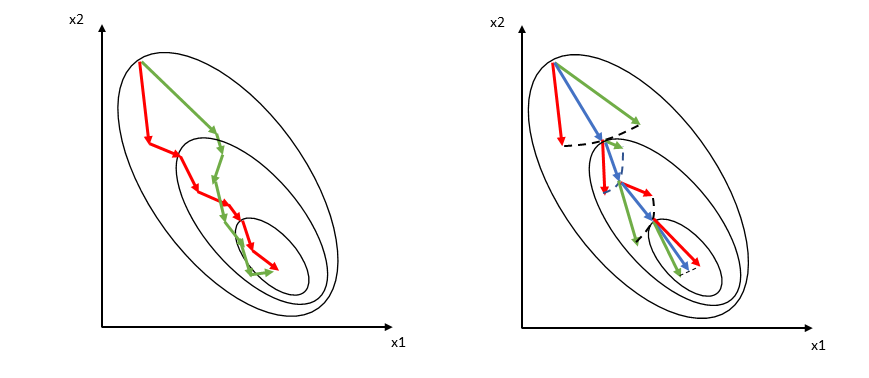} 
		\caption{ A sketch for gradient. With multiple loss functions utilized, more choices for gradients are provided, as compared in \textbf{(right)} and \textbf{(left)}. }
		\label{fig:optimization}
\end{figure}

To be more clear, we will investigate an intuitive interpretation for our nonlinear collaborative scheme. We also take $N=2$ as an example and utilize the adaptive rule, with a corresponding sketch for gradient descent given in Figure \ref{fig:optimization}.

Advantage over the single-loss scheme is apparent: Assuming that the optimal gradient which leads to the global optimization solutions can be expressed as $\nabla \widehat{\mathcal{L}}$, while neither $\mathcal{L}_1(x)$ nor $\mathcal{L}_2(x)$ can lead to $\nabla \widehat{\mathcal{L}}$ directly. With multiple loss function utilized, we can find a \textit{data-driven gradient} which combines the advantages from different loss functions, instead of finding the gradient for $\mathcal{L}_1(x)$ or $\mathcal{L}_2(x)$ separately. It helps to avoid the saddle point and poor-quality local points to some extent.

Advantages over the multi-loss scheme can be stated from generalization perspective.  Supposing that $c$ denotes the real target on the unseen test set in our task, $H_j$ ($j=1,2,\cdots$) refers to \textit{hypothesis} and it severs as \textit{ a surrogate} during training, Supposing a distribution $\mu$, the \textit{risk} or \textit{difference} between $c$ and $H_j$ can be measure by 
\[ \epsilon(h)=\mu\left( H_j(X) \neq c(X) \right)=\mu(c \Delta H_j).\]
In learning task, $\mu$ serves equivalently as test accuracy. Now we can explain why nonlinear collaboration make better sense on generalization. Supposing that linear and nonlinear collaboration can be expressed as $H_1+H_2$ and $\left( H_1^p + H_2^p \right)^{\frac{1}{p}}$, respectively, with $p>1$. For real target $c$, we compare the corresponding risks are respectively $\mu \left[ c \Delta \left(  H_1 + H_2\right) \right] $ and $\mu \left[ c \Delta \left( H_1^p + H_2^P\right)^{\frac{1}{p}} \right]$. Thus, we obtain that $\mu \left[ c \Delta \left(  H_1 + H_2\right) \right] > \mu \left[ c \Delta \left( H_1^p + H_2^p\right)^{\frac{1}{p}} \right]$ with a large probability $\frac{H_1^p + H_2^p}{\left(  H_1 + H_2\right)^p}$. 


Owing to above, we argue that our proposed algorithm framework have the following advantages: better gradient choice over single-loss scheme, and better generalization (i.e., lower risk) over multi-loss scheme. Rigorous mathematical demonstration will be given in the next Sections~\ref{sec:alg} and~\ref{sec:theo} and experiments in~\ref{sec:exp}.


\section{Mathematical Demonstration For Collaboration and Nonlinear}\label{sec:alg}

In this section, we mainly demonstrate the advantage of nonlinear collaboration based on mathematical computation, with a line using the classical single-loss scheme. Supposing the case $N=2$, the objective and gradient functions in the three schemes (i.e., single-loss, multi-loss and nonlinear collaborative schemes) can be given as \footnote{According to the method to uniform dimension as shown in Appendix A in our supplementary, here we assume that for $\forall m$, $\mathcal{L}_m(x)$ has the same dimension and they can be added to compute. }

\begin{itemize}
\item[(i)] \textit{Single-loss scheme}: 
\begin{align*}
& \mathcal{L}_{single}(x) = \mathcal{L}_1(x),\ \ \ \nabla \mathcal{L}_{single}(x) = \nabla \mathcal{L}_1(x), \\
\mathrm{or} \  & \mathcal{L}_{single}(x) = \mathcal{L}_2(x),\ \ \ \nabla \mathcal{L}_{single}(x) = \nabla \mathcal{L}_2(x). 
\end{align*}

\item[(ii)] \textit{Multi-loss scheme}: 
\begin{align*}
& \mathcal{L}_{multi}(x) = \beta_1 \mathcal{L}_1(x)+\beta_2 \mathcal{L}_2(x),\\ 
& \nabla \mathcal{L}_{multi}(x) =\beta_1 \nabla \mathcal{L}_1(x)+ \beta_2 \nabla \mathcal{L}_2(x). 
\end{align*}

\item[(iii)] \textit{Nonlinear collaborative scheme}:
\begin{align*}
& \mathcal{L}_{non}(x)  = \left[\mathcal{L}_1(x)^p+\mathcal{L}_2(x)^p\right]^{\frac{1}{p}},\\
& \nabla \mathcal{L}_{non}(x)  =  \left[ \beta_1 \mathcal{L}_1(x)^p+ \beta_2 \mathcal{L}_2(x)^p\right]^{\frac{1}{p}-1} \times \nonumber \\
& \ \ \ \left[\beta_1 \mathcal{L}_1(x)^{p-1} \nabla \mathcal{L}_1(x)+\beta_2 \mathcal{L}_2(x)^{p-1} \nabla \mathcal{L}_2(x) \right].
\end{align*}
\end{itemize}
Next, we mainly focus on the difference between (i) and (ii) or (iii) in subsection \ref{subsec:col} for the advantage of collaboration, while the comparison between (ii) and (iii) will be given in subsection \ref{subsec:non} for the advantage of nonlinear form. We will prove that the nonlinear collaborative scheme allows for smaller KL distance with respect to optimal solutions. 
Especially, we also demonstrate that the advantage of our proposed scheme can be strengthened with nonlinearity (i.e., $p$) increasing, under certain constraints. 

\subsection{Demonstration for Collaboration} \label{subsec:col}

Next, from Kullback-Leibler (KL) divergence perspective, we will demonstrate the necessity of collaboration in this subsection, showing one main difference between our proposed scheme and multi-loss one. Generally, as $Q$ and $P$ refer to the probability measures defined on $\mathbb{R}^n$, and $Q$ is continuous with respect to $P$, the KL divergence from $Q$ to $P$ can be defined as
\begin{align*}
D_{KL}(P(x) \Vert Q(x)) & = \int_x P(x) \log \frac{P(x)}{Q(x)} \mathrm{d} x, 
\end{align*}
with $P(x)$ and $Q(x)$ continuous random variables. 
Assuming the Boltzmann distribution corresponding to the optimal solutions can be expressed es $\mathcal{P}_{opt}$, we can investigate the corresponding Boltzmann distribution and KL distance in the following three schemes:   

\begin{itemize}
\item Single-loss scheme: After transforming the loss function as Boltzmann distribution $\mathcal{P}_{single} \propto \mathrm{exp}\left[ -\mathcal{L}_{single}(x) \right] $, the KL distance from optimal solutions to objective function is positively related to $D_{KL,single} \propto \int_x P_{opt}(x) \log \frac{P_{opt}(x)}{\mathrm{exp}\left[ -\mathcal{L}_{single}(x) \right] } \mathrm{d} x$.  

\item Multi-loss scheme (or called: multi-loss scheme): The corresponding Boltzmann distribution refers to $\mathcal{P}_{multi} \propto \mathrm{exp}\left[ -\mathcal{L}_{multi}(x) \right]$, and the KL divergence from optimal solutions to objective function satisfies that $D_{KL,multi} \propto  \int_x P_{opt}(x) \log \frac{P_{opt}(x)}{\mathrm{exp}\left[ -\mathcal{L}_{multi}(x) \right]} \mathrm{d} x$. 

\item Nonlinear collaborative scheme: The corresponding Boltzmann distribution is $\mathcal{P}_{non} \propto \mathrm{exp}\left[- \mathcal{L}_{non}(x) \right]$, while the KL divergence from optimal solutions to objective function meets that $D_{KL,non} \propto \int_x P_{opt}(x) \log \frac{P_{opt}(x)}{\mathrm{exp}\left[- \mathcal{L}_{non}(x) \right] } \mathrm{d} x$. 
\end{itemize}
Thus, we have $D_{KL,multi} < D_{KL,single}$ and $D_{KL,non} < D_{KL,single}$ owing to 
$\mathrm{exp}\left[ -\mathcal{L}_{multi}(x) \right] < \mathrm{exp}\left[ -\mathcal{L}_{single}(x) \right]$ and $\mathrm{exp}\left[ -\mathcal{L}_{non}(x) \right] < \mathrm{exp}\left[ -\mathcal{L}_{single}(x) \right]$. It means that \textit{the collaborative scheme}, including both linear and nonlinear ones, \textit{leads to smaller KL distance}.


\subsection{Demonstration for Nonlinear Manner}\label{subsec:non}

In this subsection, we will verify another advantage for the nonlinear collaborative scheme from nonlinear perspective: 

\subsubsection{KL Distance Perspective}

After the KL distance in nonlinear scheme is viewed as a function with respect to $p$, we have 
\begin{align*}
\frac{\partial D_{KL,non}}{\partial p} 
= & \int - \frac{P_{opt}}{e^{-\mathcal{L}_{non}(x)}} \frac{\partial e^{-\mathcal{L}_{non}(x)}}{\partial p} \mathrm{d} x \\
= & \int P_{opt}  \left( \beta_1 \mathcal{L}_1^p + \beta_2 \mathcal{L}_2^p \right)^{\frac{1}{p}-1} \\
& \times \left( \beta_1 \mathcal{L}_1^p \log \mathcal{L}_1 + \beta_2 \mathcal{L}_2^p \log \mathcal{L}_2 \right) \mathrm{d} x.
\end{align*}
Since $0 < \mathcal{L}_1,\mathcal{L}_2 <1$ practically, thus, we have $\log \mathcal{L}_1 <0$ and $\log \mathcal{L}_2 <0$. 
Hence, $\frac{\partial D_{KL,non}}{\partial p} < 0$, denoting that the \textit{KL distance from optimal solutions to objective functions decreases with $p$ increasing}. We can expect that with the appropriate loss functions for given deep learning task, the stronger nonlinearity allows for the smaller KL distance.

\subsubsection{Gradient Perspective} 

Firstly, we regard $\nabla \mathcal{L}_{non}$ as a derivable function of $p$, which yields that 
\begin{align*}
 \frac{\partial \nabla \mathcal{L}_{non}}{\partial p} = & M(p)^{\frac{1}{p}-1} \times \bigg[ -\frac{1}{p^2} \log \left( \beta_1 \mathcal{L}_1^p+ \beta_2 \mathcal{L}_2^p \right) \\
&+ \left( \frac{1}{p}-1 \right) \frac{\beta_1 \mathcal{L}_1^p \log \mathcal{L}_1 + \beta_2 \mathcal{L}_2^p \log \mathcal{L}_2}{\beta_1 \mathcal{L}_1^p+ \beta_2 \mathcal{L}_2^p }  \bigg] \\
& \times \bigg[\beta_1 \mathcal{L}_1^{p-1} \nabla \mathcal{L}_1 +\beta_2 \mathcal{L}_2^{p-1} \nabla \mathcal{L}_2 \bigg]+ M(p)^{\frac{1}{p}-1}  \\
& \times \bigg[ \beta_1 \mathcal{L}_1^{p-1} \log \mathcal{L}_1 \nabla \mathcal{L} 
 + \beta_2 \mathcal{L}_2^{p-1} \log \mathcal{L}_2 \nabla \mathcal{L}_2  \bigg],
\end{align*}
with $M(p)= \beta_1 \mathcal{L}_1^p+ \beta_2 \mathcal{L}_2^p$. Supposing that $p>1$, $0<\mathcal{L}_1, \mathcal{L}_2<1$, $0<M(p)<1$ and $\nabla \mathcal{L}_1, \nabla \mathcal{L}_2 <0$, we have 
\begin{align*}
& -\frac{1}{p^2} \log \left( \beta_1 \mathcal{L}_1^p+ \beta_2 \mathcal{L}_2^p \right) >0, \\
& \frac{\beta_1 \mathcal{L}_1^p \log \mathcal{L}_1 + \beta_2 \mathcal{L}_2^p \log \mathcal{L}_2}{\beta_1 \mathcal{L}_1^p+ \beta_2 \mathcal{L}_2^p } <0, \\
& \beta_1 \mathcal{L}_1^{p-1} \nabla \mathcal{L}_1 +\beta_2 \mathcal{L}_2^{p-1} \nabla \mathcal{L}_2 <0, \\
& \beta_1 \mathcal{L}_1^{p-1} \log \mathcal{L}_1 \nabla \mathcal{L} 
 + \beta_2 \mathcal{L}_2^{p-1} \log \mathcal{L}_2 \nabla \mathcal{L}_2 >0, 
\end{align*}
denoting that $\frac{\partial \nabla \mathcal{L}_{non}}{\partial p} >0$. It means that when the gradient of $\mathcal{L}_{non}$ is viewed as a derivable function of $p$, $\mathcal{L}_{non}$ is a monotone increasing function with the variable as $p$, implying that when $p>1$, $\nabla \mathcal{L}_{non} > \nabla \mathcal{L}_{multi}$. 
Owing to  
\begin{align*}
\nabla^2 \mathcal{L}_{non}(x) 
& =  \frac{p}{1-p} [\beta_1 \mathcal{L}_1(x)^p + \beta_2 \mathcal{L}_2(x)^p]^{\frac{1}{p}-2} \times \\
& \left[\beta_1 \mathcal{L}_1(x)^{p-1} \frac{\partial \mathcal{L}_1(x)}{\partial x} +\beta_2 \mathcal{L}_2(x)^{p-1} \frac{\partial \mathcal{L}_2(x)}{\partial x}  \right]^2 \\
& + [\beta_1 \mathcal{L}_1(x)^p + \beta_2 \mathcal{L}_2(x)^p]^{\frac{1}{p}-1} \times \\
& \bigg\{ \beta_1 (p-1) \mathcal{L}_1(x)^{p-2} \left[\nabla \mathcal{L}_1(x) \right]^2 +\beta_1 \mathcal{L}_1(x)^{p-1}  \\
& \times \nabla^2 \mathcal{L}_1(x)+ \beta_2 (p-1) \mathcal{L}_2(x)^{p-2} \times \\
& \left[ \nabla \mathcal{L}_2(x) \right]^2 +  
 \beta_2 \mathcal{L}_2(x)^{p-1}\nabla^2 \mathcal{L}_2(x) \bigg\}, 
\end{align*}
we have 
\begin{align*}
\frac{\partial \nabla^2 \mathcal{L}_{non}}{\partial p} = \frac{\partial A}{\partial p}+ \frac{\partial B}{\partial p},
\end{align*}
where
\begin{align*}
\frac{\partial A}{\partial p} & = \frac{1}{(1-p)^2} M(p)^{\frac{1}{p}-2} \times \bigg[\beta_1 \mathcal{L}_1(x)^{p-1} \nabla \mathcal{L}_1(x) \\
& + \beta_2 \mathcal{L}_2(x)^{p-1} \nabla \mathcal{L}_2(x)  \bigg]^2 +\frac{p}{1-p} M(p)^{\frac{1}{p}-2}  \\
& \times \bigg[ -\frac{1}{p^2} \log M(p) + \left(\frac{1}{p}-2 \right) \frac{1}{M(p)} \\
& \times (\beta_1 \mathcal{L}_1^p \log \mathcal{L}_1+ \beta_2 \mathcal{L}_2^p \log \mathcal{L}_2)\bigg] \\
& \times \left[\beta_1 \mathcal{L}_1^{p-1} \nabla \mathcal{L}_1 +\beta_2 \mathcal{L}_2^{p-1} \frac{\partial \mathcal{L}_2}{\partial x}  \right]^2 + \frac{2 p}{1-p} \\
& \times M(p)^{\frac{1}{p}-2} 
\left[\beta_1 \mathcal{L}_1^{p-1} \nabla \mathcal{L}_1 +\beta_2 \mathcal{L}_2^{p-1} \nabla \mathcal{L}_2  \right] \\
& \times \left[\beta_1 \mathcal{L}_1^{p-1} \log \mathcal{L}_1 \nabla \mathcal{L}_1 +\beta_2 \mathcal{L}_2^{p-1} \log \mathcal{L}_2 \nabla \mathcal{L}_2  \right]
\end{align*}
and 
\begin{align*}
\frac{\partial B}{\partial p} & = \bigg\{ \beta_1 (p-1) \mathcal{L}_1(x)^{p-2} \left[\nabla \mathcal{L}_1(x) \right]^2 +\beta_1 \mathcal{L}_1(x)^{p-1}  \\
& \times \nabla^2 \mathcal{L}_1(x)+ \beta_2 (p-1) \mathcal{L}_2(x)^{p-2}  \\
& \times \left[ \nabla \mathcal{L}_2(x) \right]^2 +  
 \beta_2 \mathcal{L}_2(x)^{p-1}\nabla^2 \mathcal{L}_2(x) \bigg\} \\
& \times M(p)^{\frac{1}{p}-1} \bigg[ -\frac{1}{p^2} \log M(p)+\left(\frac{1}{p}-1 \right) \\
& \times \frac{\beta_1 \mathcal{L}_1^p \log \mathcal{L}_1 + \beta_2 \mathcal{L}_2^p \log \mathcal{L}_2}{M(p)} \bigg] + M(p)^{\frac{1}{p}-1} \\
& \times \bigg\{ \beta_1 \left(\nabla \mathcal{L}_1 \right)^2 \left[ \mathcal{L}_1^{p-2}+ (p-1) \mathcal{L}_1^{p-2} \log \mathcal{L}_1 \right] \\
& +\beta_1 \mathcal{L}_1^{p-1} \log \mathcal{L}_1 \nabla^2 \mathcal{L}_1 + \beta_2 \mathcal{L}_2^{p-1} \log \mathcal{L}_2 \nabla^2 \mathcal{L}_2 \\
& + \beta_2 \left(\nabla \mathcal{L}_2 \right)^2 \left[ \mathcal{L}_2^{p-2}+ (p-1) \mathcal{L}_2^{p-2} \log \mathcal{L}_2 \right]  \bigg\}.
\end{align*}
Hereby, $\nabla^2 \mathcal{L}_{non}$ is also viewed as a derivable function of $p$, and in order to simplify the expression, we set $M(p)=\beta_1 \mathcal{L}_1(x)^p + \beta_2 \mathcal{L}_2(x)^p$, $A = \frac{p}{1-p} [\beta_1 \mathcal{L}_1(x)^p + \beta_2 \mathcal{L}_2(x)^p]^{\frac{1}{p}-2} \left[\beta_1 \mathcal{L}_1(x)^{p-1} \frac{\partial \mathcal{L}_1(x)}{\partial x} +\beta_2 \mathcal{L}_2(x)^{p-1} \frac{\partial \mathcal{L}_2(x)}{\partial x}  \right]^2$ as well as $B=\nabla^2 \mathcal{L}_{non}(x)-A$. 

Clearly, when $p>1$, $0<M(p)<1$, $\nabla \mathcal{L}_1, \nabla \mathcal{L}_2<0$, $\nabla^2 \mathcal{L}_1, \nabla^2 \mathcal{L}_2 <0$~\footnote{Such assumption is apparent in practice: the change of objective function slows down step by step, implying that the second-order derivative is negative. }, we have $\frac{\partial A}{\partial p}>0$, $\frac{\partial B}{\partial p}>0$ under the conditions
\begin{eqnarray} \label{eq:constraint_p}
&& (p-1) \left(\nabla \mathcal{L}_j \right)^2   
 + \mathcal{L}_j  \nabla^2 \mathcal{L}_j >0,\ \ j=1,2 
\end{eqnarray}
which means that the advantage of our nonlinear scheme can be strengthened when $p>1$ under certain constraints.

In truth, from the statistical view, we can go deeper based on Constraint (\ref{eq:constraint_p}). Accounting for the constraint, we can obtain the critical conditions for optimization: $p=1- \frac{\mathcal{L}_j \nabla^2 \mathcal{L}_j}{\left( \nabla \mathcal{L}_j \right)^2}$ and $\mathcal{L}_j(x)=c_{2j} (px-c_{1j})^{\frac{1}{p}}$, where $i,j=1,2$, $c_{ij}$ is an integral constant. Due to our assumptions $0<\mathcal{L}_j<1$ and $\nabla^2 \mathcal{L}_j<0$, we get $\frac{\mathcal{L}_j \nabla^2 \mathcal{L}_j}{\left( \nabla \mathcal{L}_j \right)^2}<0$, leading to a \textit{phase transition} when $p$ increases from $p<1- \frac{\mathcal{L}_1 \nabla^2 \mathcal{L}_1}{\left( \nabla \mathcal{L}_1 \right)^2}$ to $p>1- \frac{\mathcal{L}_1 \nabla^2 \mathcal{L}_1}{\left( \nabla \mathcal{L}_1 \right)^2}$ in our nonlinear collaborative scheme.

Therefore, from KL divergence and gradient perspectives, we can conclude the advantage of nonlinear collaborative scheme based on mathematical computation:

\begin{itemize}
\item collaborative scheme leads to smaller KL distance with respect to optimal solutions. 

\item collaborative scheme allows for data-driven choice during stochastic gradient descent. 

\item nonlinear collaboration gives rise to faster convergence, comparing with the linear one, when $\nabla \mathcal{L}_{non} > \nabla \mathcal{L}_{multi}$. 

\item advantage of nonlinear collaborative scheme can be strengthened with $p$ increasing under appropriate constraint, especially when $p > 1- \frac{\mathcal{L}_j \nabla^2 \mathcal{L}_j}{\left( \nabla \mathcal{L}_j \right)^2}$.
\end{itemize}
Owing to the above, we can expect that the new proposed nonlinear collaborative scheme can perform superior to the classical single-loss or multi-loss (i.e., linear) scheme in practice. 

\section{Generalization Ability and PAC-Bayes Bound}\label{sec:theo}

In this section, we will present our another contribution, apart from the advantage for optimization, on the connection between minimizing our objective function (\ref{eq:non_obj}) and minimizing a PAC-Bayes bound through two steps.  

\subsection{From Loss Function to Generalized Entropy}
 
Firstly, let us investigate the connection between minimizing loss function (\ref{eq:non_obj}) and maximizing a generalized entropy.

Given a loss function $f(x)$, its corresponding Gibbs distribution is $P(x,\beta)=\mathcal{Z}_{\beta}^{-1} \mathrm{exp}\left[-\beta f(x) \right]$, with $\beta$ as inverse temperature. 
As $\beta \rightarrow \infty$, such Gibbs distribution concentrates on the global minimum of $f(x)$, assuming as 
\[ x^{*} = \mathrm{argmin}_x f(x), \]
which establishes the \textit{connection between the Gibbs distribution and a generic optimization problem}. 
Therefore, for each separate loss function $\mathcal{L}_m(x)$ ($m=1,2,\cdots,N$) and our nonlinear collaborative objective function (\ref{eq:non_obj}), the corresponding Gibbs distributions are
\begin{align}\label{eq:gibbs_non}
& P_m(x,\beta)= \mathcal{Z}^{-1}_{\beta_m} \mathrm{exp} \left(-\beta_m \mathcal{L}_m(x) \right), \\
& P_{non}(x,\beta) = \mathcal{Z}_{\beta}^{-1} \mathrm{exp} \left[ -\left( \sum_{m=1}^N \beta_m \mathcal{L}_m(x)^p \right)^{\frac{1}{p}} \right]. 
\end{align}
Let pick $\beta_m \in \{ \beta_1,\beta_2,\cdots,\beta_N \}$. Apparently, when $\beta_j \ (j \neq m) \rightarrow 0$,  $\mathcal{L}_m(x)$ in the exponent dominates and the new distribution is similar to the typical Gibbs distribution. 
Thus, we will set $\sum\limits_{m=1}^N \beta_m=1$ because $N-1$ variables can afford us to control on Gibbs distribution sufficiently. Then, we can define a \textit{generalized entropy} as 
\begin{align}\label{eq:non_entropy}
& S_{non}(x,\beta)  = \log \mathcal{Z}_{\beta} \nonumber \\
= & \log \int_{x'} \mathrm{exp} \left[ -\left( \sum_{m=1}^N \beta_m \mathcal{L}_m(x)^p \right)^{\frac{1}{p}} \right] \mathrm{d}x^{'}.  
\end{align}
Obviously, minimizing (\ref{eq:non_obj}) equals to maximizing (\ref{eq:non_entropy}).

Specifically, when $N=2$ and $p=1$, the Gibbs distribution (\ref{eq:gibbs_non}) and generalized entropy (\ref{eq:non_entropy}) can be degenerated into 
\begin{align*}
	& P_{non}(x,\beta_1) =   \mathcal{Z}_{\beta_1,\beta_2}^{-1} \mathrm{exp} \left[ - \beta_1 \mathcal{L}_1(x) - (1-\beta_1) \mathcal{L}_2(x)  \right],  \\
	& S_{non}(x,\beta_1) =  \log \int_{x'} \mathrm{exp} \left[ -\beta_1 \mathcal{L}_1(x) - (1-\beta_1) \mathcal{L}_2(x)  \right] \mathrm{d}x^{'}.
\end{align*} 
Assume that $\mathcal{L}_1(x)$ refers to MSE, $\mathcal{L}_2(x)$ is another loss function like CE or JSD. The generalized entropy can be rewritten as 
\begin{align*}
-S_{non}(x,\beta) =  -\log \int_{x'} & \mathrm{exp} -\bigg[ \beta_1 \Vert f(x)-f(x') \Vert_2^{2} \\
& + (1-\beta_1) \mathcal{L}_2(x) \bigg] \mathrm{d}x^{'}.
\end{align*}
with $f(x)$ and $f(x')$ respectively as the real output and corresponding label, $x$ refers to the weights corresponding to learning, $x^{'}$ is a dummy variable corresponding to the original weight related to label. In practice, the nonlinear part of $\Vert f(x)-f(x^{'}) \Vert_2^2$ can be approximated by $\xi_i \Vert x-x^{'} \Vert_2^2$ based on Taylor expansion, where $\xi_i \in \Xi$ denotes the sampled mini-batch. Clearly, the generalized Gibbs distribution and entropy can respectively be degenerated to the modified distribution and local entropy in \cite{entropy_sgd}, which also guarantees the robustness of our nonlinear collaborative scheme. 


\subsection{From Optimizing Generalized Entropy To Minimizing PAC-Bayes Bound}

Secondly, we will demonstrate that maximizing a generalized entropy equals to minimizing a PAC-Bayes bound, and we prove that our nonlinear collaborative scheme allows for a tighter bound comparing with the linear one. Before this, we first introduce the fundamental notion and theorem.

\textbf{PAC-Bayes Bounds\cite{catoni2007pac,langford2002quantitatively}}. PAC-Bayes bounds are a generalization of the Occam’s razor bound for algorithms which output a distribution over classifiers, rather than just a single classifier.

\textbf{Lemma~1} (Property Lemma \cite{vapnik2015uniform}). PAC-Bayes bounds are much tighter than most common VC-related approaches on continuous classifier spaces, shown by application to stochastic neural networks. 


\textbf{Theorem~1} (Linear PAC-Bayes Bounds (\cite{catoni2007pac})). Fix $\lambda > \frac{1}{2}$ and assume that the loss function takes values in an interval of length $\mathcal{L}_{max}$. For every $\delta>0$, $m \in \mathbb{N}$, distribution $\mathcal{D}$ on $\mathbb{R}^k \times K$, where $K$ denotes a finite discrete set of labels, and distribution $P$ on $\mathbb{R}^p$. For $\forall Q$, we have  
\begin{align}\label{eq:linear_pac}
\mathbb{P}_{S \sim \mathcal{D}^m} & \bigg\{ \left(1-\frac{1}{2 \lambda} \right) R_{\mathcal{D}}(Q) \leq \hat{R}_S(Q) + \nonumber \\
& \frac{\lambda \mathcal{L}_{max}}{m} \bigg[ D_{KL}(Q \Vert P)+\log \frac{1}{\delta} \bigg] \bigg\} \geq 1-\delta.
\end{align} 

Using the same assumption as Sec. \ref{sec:alg} and Expression (\ref{eq:emp_risk}), we can rewrite the generalized entropy (\ref{eq:non_entropy}) as
\begin{align}\label{eq:non_p2}
- S_{non}(x,\beta) = & -\log \int_{x'} \mathrm{exp} - \bigg\{  \beta_1 \Vert f(x)-f(x') \Vert_2^{2p} + \nonumber \\
& (1-\beta_1) \left[ \widehat{R}_{S}(Q) \right]^p \bigg\}^{\frac{1}{p}} \mathrm{d}x^{'},
\end{align}
where $\frac{1}{m}\sum_{i=1}^m \mathbb{E}[ \ell(x^{'},z_i)]=\mathcal{L}_2(x)$ denotes another objective function, $x^{'} \sim Q$.
Enlightened by Theorem~$3.1$ in \cite{sgd_pac2017entropy}, we have the following corollary:

\textbf{Corollary 1:} (\textit{Maximizing generalized entropy optimizes a PAC-Bayes prior}) Maximizing the generalized entropy (\ref{eq:non_entropy}) equals to minimize a PAC-Bayes bound in Theorem 2, i.e., Inequation (\ref{eq:linear_pac}), with a prior related to $P$. Here, $P$ is a prior depends on samples/observation $S$,  $\lambda < \frac{1}{2}$, and $L_{max}$, $\delta$ and $\mathcal{D}$ have the same definitions as Theorem $2$. 

$Proof.$ Let $S \sim \mathcal{D}^m$ denote the observation. Clearly, essence of optimizing $-S_{non}(x,\beta)$ in (\ref{eq:non_p2}) equals to obtain the following gradient 
\begin{align*}
2 \beta_1 p \Vert f(x)-f(x') \Vert_2^{2p-1} \langle f(x)-f(x^{'}) \rangle + \\
 p(1-\beta_1)  \left[ \widehat{R}_{S}(Q) \right]^{p-1} \nabla \widehat{R}_{S}(Q), 
\end{align*}
where we ignore the gradient of $\mathrm{e}$ exponential information and the original gradient of exponent itself which are easily computed. Key point of optimization lies within the SGLD sampling and gradient of $\widehat{R}_S(Q)$. 
To verify the PAC-Bayes bound, we need to demonstrate that for the least probability $1-\delta$, there exists 
\begin{align}\label{eq:coro_1}
\left(1-\frac{1}{2 \lambda} \right) R_{\mathcal{D}}(Q) \leq \widehat{R}_S(Q)+\frac{\lambda \mathcal{L}_{max}}{m} \left[ D_{KL}(Q \Vert P)+\log \frac{1}{\delta} \right].
\end{align}
Minimizing $R_{\mathcal{D}}(Q)$ is equivalent to minimizing the upper bound
\[ \mathrm{inf}_{Q} \ \  \frac{1}{m} \sum_{i=1}^m \mathbb{E}_{x \sim Q}[\ell(x,z_i)]+ \lambda \frac{\mathcal{L}_{max}}{m} D_{KL}(Q \Vert P). \]
Owing to $\frac{1}{m} \sum_{i=1}^m \mathbb{E}_{x \sim Q}[\ell(x,z_i)] = \mathbb{E}_{x \sim \widehat{D}} [\ell(x,z_i)]=R_{\widehat{S}}(x)$, the upper bound can be rewritten as $\mathrm{inf}_Q \ \ \mathbb{E}_{x \sim Q} [R_{\widehat{S}}(x)]+ \lambda \frac{\mathcal{L}_{max}}{m} (D_{KL}(Q \Vert P)$, i.e., $\mathrm{inf}_Q \ \ P(R_{\widehat{S}}(x))+ \lambda \frac{\mathcal{L}_{max}}{m} (D_{KL}(Q \Vert P)$. 
According to Lemma $1.1.3$ in \cite{catoni2007pac}, we have \footnote{For a measure $P$ on $\mathbb{R}^p$ and function $g: \mathbb{R}^p \rightarrow \mathbb{R}$, let $P[g]:= \int g(h) P(\mathrm{d}h)$ denotes   \textit{Expectation} and $P[g] < \infty$, $P_g$ refers to the probability measure on $\mathbb{R}^p$ \cite{sgd_pac2017entropy,langford2003pac}.} 

\begin{align*}
\log  P[\mathrm{exp} & (-r(x))]  = \mathbb{E}_{x \sim Q}[r(x)]- \\
& D_{KL}(Q \Vert P) + D_{KL}(Q \Vert P_{\mathrm{exp}(-r) })  .
\end{align*}

Combining the two expressions above, minimizing $R_D(Q)$ equals to minimizing 
\begin{align}\label{eq:inf_empirical}
\mathrm{inf}_{Q} \ \  D_{KL}(Q \Vert P_{\mathrm{exp}(-r) })  -  \log P [\mathrm{exp}(-r(x))].
\end{align}
When $Q=P[\mathrm{exp}(-r(x))]$, $D_{KL}(Q \Vert P[\mathrm{exp}(-r(x))] )=0$ reaches the minimum. Expression (\ref{eq:inf_empirical}) equals to $- \log P[\mathrm{exp}(-r(\mathbf{w}))]$. Thus, Expression (\ref{eq:coro_1}) yields that 
\begin{align*}
& \left(1-\frac{1}{2 \lambda} \right) R_{\mathcal{D}}(P[\mathrm{exp}(-r(x)) ]) \\
\leq & -  \log P[\mathrm{exp}(-r(x)) ]+ \log \frac{1}{\delta}.
\end{align*}

Therefore, it is plain to see that optimizing the generalized entropy (\ref{eq:non_p2}) equals to optimize a PAC-Bayes bound with the prior $P_{\mathrm{exp(-\widehat{R}_S(Q))}}$ which depends on the samples/observation $S$. 
$\hfill\blacksquare$

On account of Corollary~$1$, we can get:  

\begin{itemize}
\item optimizing generalized entropy, i.e., our proposed nonlinear collaboration algorithmic scheme, equals to minimize a PAC-Bayes bound with respect to the prior $P$. 

\item prior $P$ can be changed in a differentially private way\footnote{ Corresponding theory and calculation are given in Appendix D.}, considering the PAC-Bayes theorem from the priori perspective since such bound is not met when the prior depends on the sample $S$ \cite{dwork2008differential,sgd_pac2017entropy}. 

\item KL divergence between $\widehat{R}_S(Q)$ and $R_{\mathcal{D}}(Q)$ is  leveraged to replace $R_{\mathcal{D}}(Q)-\widehat{R}_S(Q)$, making up the looser bound induced by differentially private prior. 

\item different PAC-bound can be compared through comparing $D_{KL}\left(\widehat{R}_S(Q) \Vert R_{\mathcal{D}}(Q)\right)$ in different cases. 
\end{itemize}

Thus, we can obtain the new PAC-Bayes bound and advantage on generalization ability of our nonlinear collaborative scheme based on Theorem $5.4$ in \cite{sgd_pac2017entropy} and Theorem $6$ \cite{mcsherry2007mechanism} , with the demonstration shown in Appendix E:

\textbf{Corollary 2:} Under $0-1$ loss, for $\forall \delta$ and $\forall Q$, $m \in \mathbb{N}$, and $\epsilon$-differential private data-dependent prior $P_{\mathrm{exp}(S)}$, we have 
\begin{align*}
& \mathbb{P}_{S \sim \mathcal{D}^m} \bigg\{ D_{KL}\left[\widehat{R}_S(Q) \Vert R_{\mathcal{D}}(Q) \right]  \\
& \hspace{5mm} \leq  \frac{ D_{KL}(Q \Vert P_{\mathrm{exp}(S)} ) +\log 2m +2 \mathrm{max} \left\{\log \frac{3}{\delta},m \epsilon^2 \right\} }{m-1} \bigg\} \\
\geq & 1-\delta.
\end{align*}
\textbf{Corollary 3:} Comparing with the local entropy in \cite{entropy_sgd}, our proposed generalized entropy allows for a tighter PAC-Bayes bound. 

$Proof.$ Its key point is to compare $KL = D_{KL}\left[\widehat{R}_S(Q) \Vert R_{\mathcal{D}}(Q) \right]$ in different cases. Set indexes $local$ and $gene$ represent the KL divergence for the optimal solutions to different entropies, respectively: one is local entropy in \cite{entropy_sgd}, the other one is the generalized entropy in our nonlinear collaborative scheme. 

Firstly, we have $\left[\mathcal{L}_1(x)^p+\mathcal{L}_2(x)^p\right]^{\frac{1}{p}} < \mathcal{L}_1(x)+\mathcal{L}_2(x)$, which yields that the generalized entropy (\ref{eq:non_entropy}) is larger than local entropy in the context of $\ell_2$ norm. Owing to the definition, we have $KL_{local}< KL_{gene}$. Clearly, even though each PAC-Bound would be looser owing to the differential privacy, KL term which can be utilized to make up the gap enjoys a larger value in our \textit{nonlinear collaboration}, resulting in a relatively \textit{tighter} PAC-Bayes bound. 
$\hfill\blacksquare$

Thanks to \textit{differential privacy} \cite{dwork2008differential}, we do not need to worry about whether the prior $P$ in our PAC-Bayes bound depends on samples/observations $S$ or not. To make the paper more completed, we introduce differential privacy briefly in Appendix B. Furthermore, another view on generalization ability of our nonlinear collaborative scheme is given in Appendix C, showing the advantage of our nonlinear collaborative scheme intuitively and graphically.

\section{Generalization Ability From Fourier Analysis} \label{sec:fourier}

Briefly speaking, we advocate viewing network training as an optimization problem, and our proposed scheme contributes to better surrogate for $0-1$ loss function and tighter PAC-Bayes bound. In this way, it becomes clear that with nonlinearity increasing, both learning and generalization can be improved. It is worth mentioning that several numerical evidence can support this view recently, e.g., \textit{sharpness} \cite{pac_sharp2016large} and \textit{spectral analysis} \cite{pac_spec2018understanding,pac_spec2018training}.   
Based on these, we aim to demonstrate our proposed scheme further on generalization ability.

\subsection{Sharpness and Generalization}

Several experiments verify that due to the inherent noise in gradient estimation, sharp minimizers of the training and testing functions lead to the generalization drop in the large-batch regime \cite{pac_sharp2016large}. Given this, sharpness, 
\begin{align*}
\zeta_{\alpha} & = \frac{\mathrm{max}_{\mid \nu_i \mid \leq \alpha (\mid \mathbf{w}_i \mid +1)} \widehat{R}_S(\mathbf{w}+\nu) - \widehat{R}_S(\mathbf{w})}{1+ \widehat{R}_S(\mathbf{w}) } \\
& \simeq \mathrm{max}_{\mid \nu_i \mid \leq \alpha (\mid \mathbf{w}_i \mid +1)} \widehat{R}_S(\mathbf{w}+\nu) - \widehat{R}_S(\mathbf{w}),
\end{align*}
is designed to measure the generalization ability and corresponds to robustness to adversarial perturbations on the parameter space \cite{pac_sharp2016large,pac_gene2017exploring}. Hereby, $\alpha$ is a hyperparameter, $\nu$ is a random variable, and $\mathbf{w}$ denotes a single predictor learned from the training set.

In the context of PAC-Bayes framework, with the prior $P$ and probability at least $1-\delta$ over the draw of training data, \textit{expected sharpness} can be bounded as \cite{pac_gene2017exploring,pac_gene2017computing}
\begin{align*}
& \mathbb{E}_{\nu}\left[\widehat{R}_S(\mathbf{w}+\nu) \right] \\
\geq & \mathbb{E}\left[ R_D(Q) \right] - 4 \sqrt{\frac{D_{KL}(\mathbf{w}+\nu \Vert P)+\log \frac{2m}{\delta} }{m}}, 
\end{align*}
which equals that 
\begin{align*}
& \zeta_{\alpha} \simeq \mathbb{E}_{\nu}\left[\widehat{R}_S(\mathbf{w}+\nu) \right] - \mathbb{E}_{\nu}\left[\widehat{R}_S(\mathbf{w}) \right] \\
\geq & \mathbb{E}\left[ R_D(Q) \right]-\mathbb{E}_{\nu}\left[\widehat{R}_S(\mathbf{w}) \right] - 4 \sqrt{\frac{D_{KL}(\mathbf{w}+\nu \Vert P)+\log \frac{2m}{\delta} }{m}}. 
\end{align*}
Obviously, the lower bound of expected sharpness, the more likely it is that algorithmic scheme leads to the better generalization ability.

Comparing the nonlinear collaborative, multi-loss and single-loss schemes, we can easily find that $e^{-\left(\mathcal{L}_1^p+\mathcal{L}_2^p \right)^{\frac{1}{p}}} < e^{-\mathcal{L}_1}$ and $e^{-\left(\mathcal{L}_1^p+\mathcal{L}_2^p \right)^{\frac{1}{p}}} < e^{-\mathcal{L}_2}$. Apparently, it leads to a larger KL divergence $D_{KL}(\mathbf{w}+\nu \Vert P)$ in nonlinear collaborative scheme, which means a smaller expected sharpness and better generalization. 
Utilizing Residual Networks and Densenet, we will compare the three algorithmic schemes on random labels in Section \ref{sec:exp}, showing the superiority on generalization ability from another perspective. 

\subsection{Spectral Analysis and Generalization} 

Utilizing Fourier analysis, it has been found that DNNs with stochastic gradient-based methods endows low-frequency components of the target function with a higher priority during the training \cite{pac_spec2018training,pac_spec2018understanding}. It means that a DNN with common settings first quickly captures the dominant low-frequency components, and then relatively slowly captures high-frequency ones \cite{pac_spec2018training,pac_spec2018understanding,pac_spec2018spectral}. In this subsection, we will discuss the nonlinear collaborative scheme in Fourier domain, showing the superiority on capture high-frequency components and generalization ability.

Firstly, we investigate a basic definition for Fourier transform: For a sigmoid function $\sigma(x)=\frac{\mathrm{e}^x}{\mathrm{e}^x+1}$, the Fourier transform for $\sigma(ax+b)$ is 
\begin{align*}
F\left[\sigma(ax+b) \right](\omega) = - \frac{i \pi}{|a|} \mathrm{exp} \left(\frac{ib \omega}{a} \right) \frac{2}{\mathrm{exp} \left(\frac{\pi \omega}{a} \right) - \mathrm{exp} \left(-\frac{\pi \omega}{a} \right)},
\end{align*}
where $a,b \in \mathbb{R}$, $\omega$ refers to the frequency, and
\[ F[\sigma(x)](\omega)= \int \sigma(x) \mathrm{exp} (-i \omega x) \mathrm{d} x. \]

We also assume that $N=2$, and $\mathcal{L}_{single}(x)$, $\mathcal{L}_{multi}(x)$ as well as $\mathcal{L}_{non}(x)$ respectively denote the objective functions for single-loss, multi-loss and nonlinear collaborative schemes, with the same forms as Section \ref{sec:alg}~\footnote{Without loss of generalization, we assume that $\mathcal{L}_{single}(x)=\mathcal{L}_1(x)$.}. Then using Fourier transform, we suppose that these objective functions can be rewritten as $\mathcal{L}(\omega)$, $\mathcal{L}_{multi}(\omega)$ as well as $\mathcal{L}_{non}(\omega)$. Obviously, $\mathcal{L}_{single}(x)=\mathcal{L}_j(x)$. Owing to $(3.13)$ in \cite{pac_spec2018understanding}, we have 

\begin{align*}
& \frac{\partial \mathcal{L}_{single}(\omega)}{\partial \Theta_{jk}}  = A_1(\omega) \mathrm{exp}\left(- \left| \frac{\pi \omega}{a_j} \right| \right) G_{jk}^0, \\
& \frac{\partial \mathcal{L}_{multi}(\omega)}{\partial \Theta_{jk}}  = \left[ \beta_1 A_1(\omega) G_{jk}^{1}+\beta_2 A_2(\omega) G_{jk}^{2} \right] \mathrm{exp}\left(- \left| \frac{\pi \omega}{a_j} \right| \right),\\
& \frac{\partial \mathcal{L}_{non}(\omega)}{\partial \Theta_{jk}}  = \left[ \beta_1 \mathcal{L}_1(\omega)^p+ \beta_2 \mathcal{L}_2(\omega)^p\right]^{\frac{1}{p}-1} \bigg[\beta_1 \mathcal{L}_1(\omega)^{p-1}  \\
& \hspace{7mm} \times \frac{\partial \mathcal{L}_1(\omega)}{\partial \omega}+ \beta_2 \mathcal{L}_2(\omega)^{p-1} \frac{\partial \mathcal{L}_2(\omega)}{\partial \omega}\bigg] \\
&\hspace{7mm}  = \mathrm{exp} \left(- \left| \frac{\pi \omega}{a_j} \right| \right) \left[ \beta_1 \mathcal{L}_1(\omega)^p+ \beta_2 \mathcal{L}_2(\omega)^p\right]^{\frac{1}{p}-1}  \\
& \hspace{7mm} \times \bigg[\beta_1 \mathcal{L}_1(\omega)^{p-1}  A_1(\omega) G_{jk}^{1} + \beta_2 \mathcal{L}_2(\omega)^{p-1} A_2(\omega) G_{jk}^{2} \bigg] 
\end{align*}
where the frequency $\omega$ is introduced by Fourier transform, $\Theta_j \triangleq \{a_j,b_j,\alpha_j\}$, $A_j(\omega)$ is a function of $\omega$, $\Theta_{jk} \in \Theta_j$, and $G_{jk}^m \ (m=0,1,2)$ refers to a function of $\Theta_j$ and $\omega$.

Apparently, $A(\omega)  \mathrm{exp} \left(- \left| \frac{\pi \omega}{a_j} \right| \right)$ dominates the behaviors of $\frac{\partial \mathcal{L}_{single}(\omega)}{\partial \Theta_{jk}}$ and $\frac{\partial \mathcal{L}_{multi}(\omega)}{\partial \Theta_{jk}}$ in DNNs, since (i) before training, the frequency $\omega$ does not converge and $a_j$ is small, and (ii) after training, $\mathrm{exp} \left(- \left| \frac{\pi \omega}{a_j} \right| \right)$ comes from the exponential decay of the activation function in the Fourier domain and the contribution of $\omega$ to the total descent amount vanishes.

Different phenomenon occurs for $\frac{\partial \mathcal{L}_{non}(\omega)}{\partial \Theta_{jk}}$: even when $\omega$ does not converge and $a_j$ is small, $\mathrm{exp} \left(- \left| \frac{\pi \omega}{a_j} \right| \right) A_j(\omega)$ cannot dominate the whole equation because of $\mathcal{L}_1(\omega)^{p-1}$ and $\mathcal{L}_2(\omega)^{p-1}$, especially when $p>1$. It means that the low-frequency components are not dominant and high-frequency ones can be captured more easily, which servers as another reason for a better generalization ability.


\section{Experiments}\label{sec:exp}

To evaluate our nonlinear collaborative scheme, we check two standard deep neural networks, Residual Networks and Densenet, on CIFAR-10 and CIFAR-100. All the comparisons are conducted by the following three schemes (without loss of generalization, we assume that $N=2$):

\begin{itemize}
	\item Single-loss scheme: The loss function is chosen CE, i.e., $\mathcal{L}(x)= L_{ce}$. 
	
	\item Multi-loss (i.e, linear) scheme: The objective function refers to $\mathcal{L}(x)=\beta_1 L_{ce}+ \beta_2 L_{mse}$. 
	
	\item Nonlinear collaborative scheme: The objective function is $\mathcal{L}(x)=\left(\beta_1 L_{ce}^p+ \beta_2 L_{mse}^{p}\right)^{\frac{1}{p}}$. 
\end{itemize}
In this section, using different levels of randomization of labels, we will explore:

\begin{enumerate}
\item[(1)] the advantage on train and validation accuracy using nonlinear collaborative algorithms with different networks and datasets. 

\item[(2)] whether our nonlinear collaborative scheme allows for more robustness in practice, as claimed in \cite{janocha2017loss,xu2016multi}.

\item[(3)] whether our new proposed scheme can lead to relatively large accuracy in the context of random labels.
\end{enumerate}

\subsection{General Experiments:}

\begin{figure}[!htbp]
		     \centering
		\includegraphics[width=0.49\linewidth]{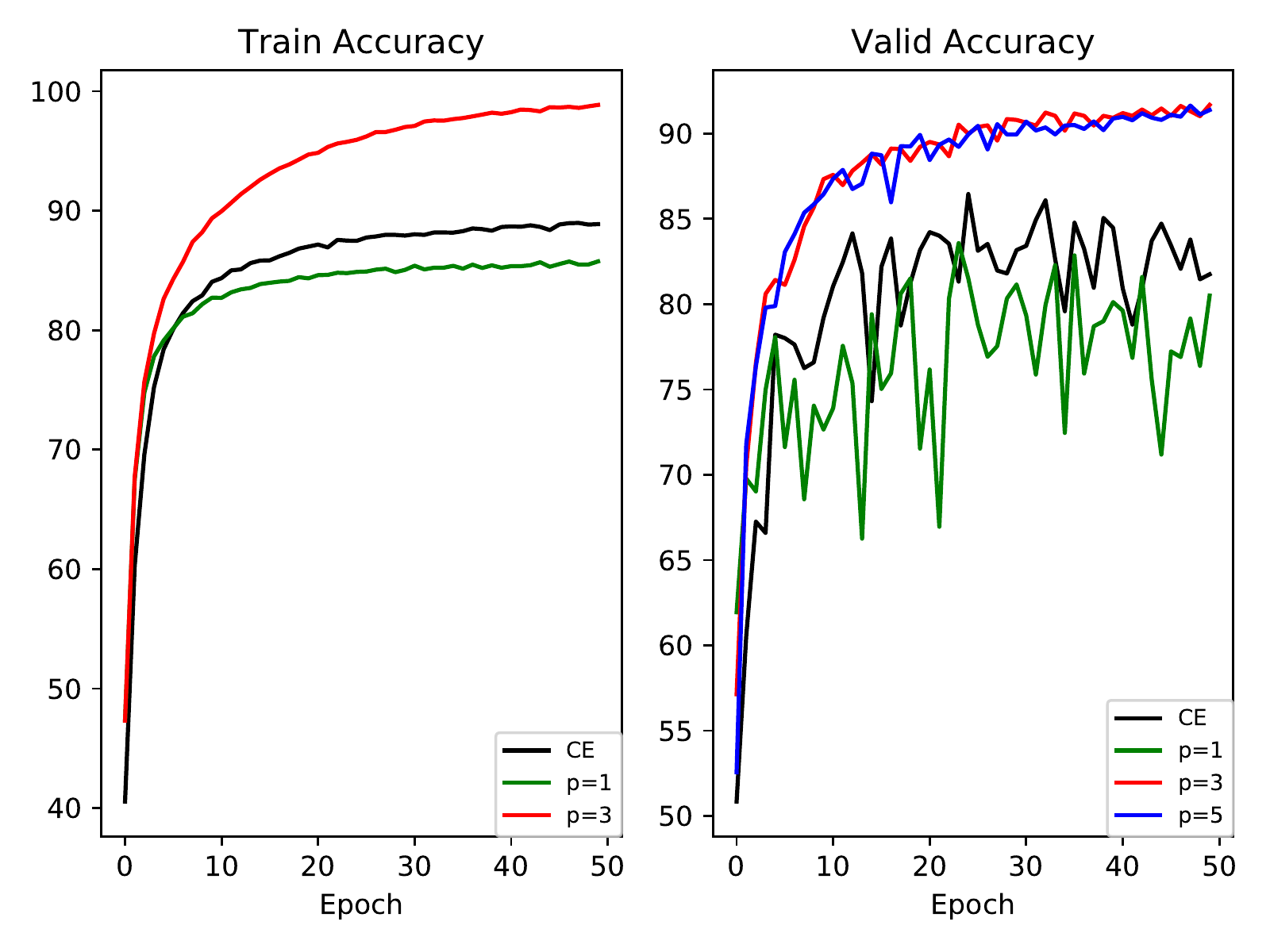} 
			\includegraphics[width=0.49\linewidth]{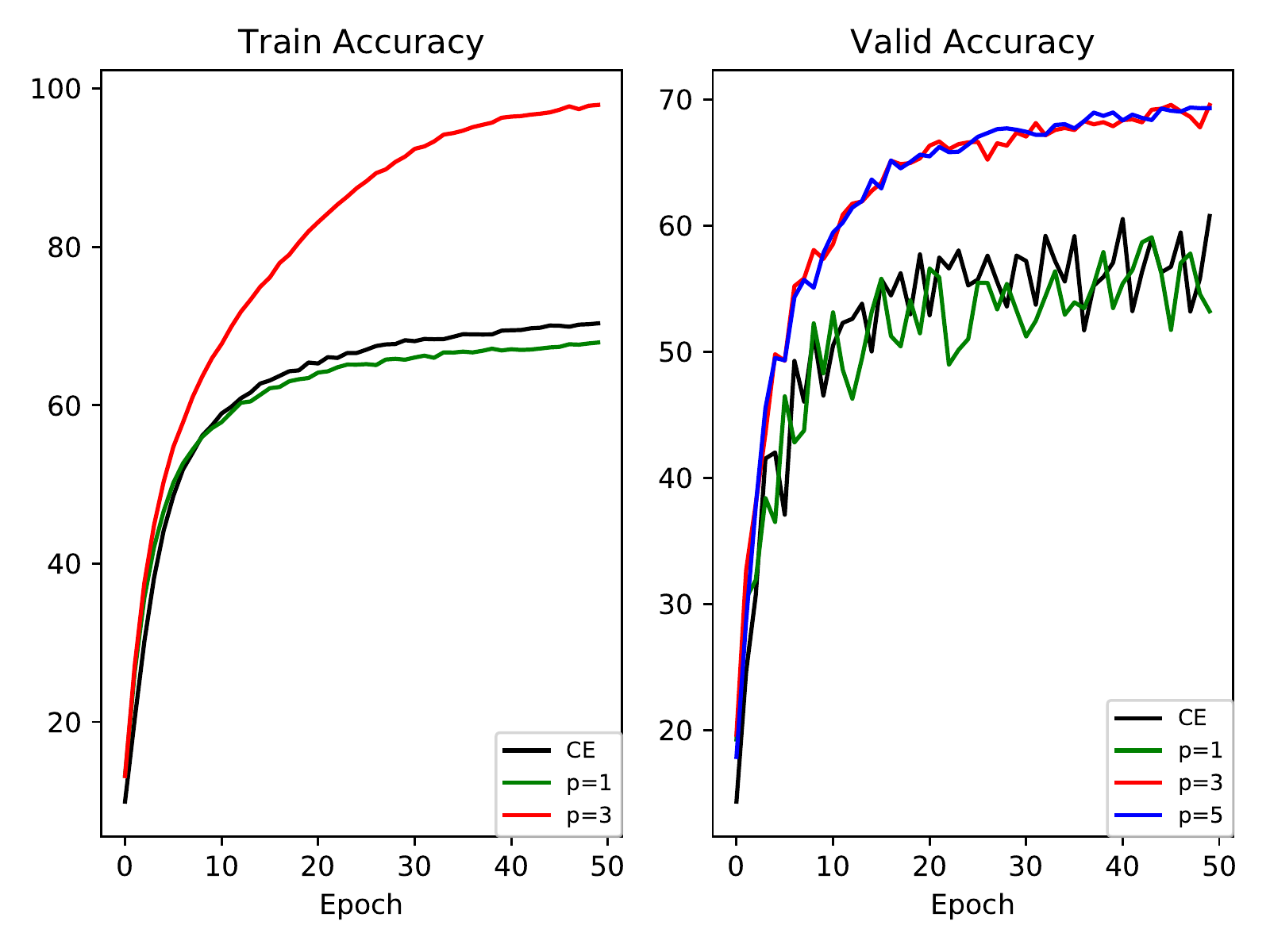}
            \includegraphics[width=0.49\linewidth]{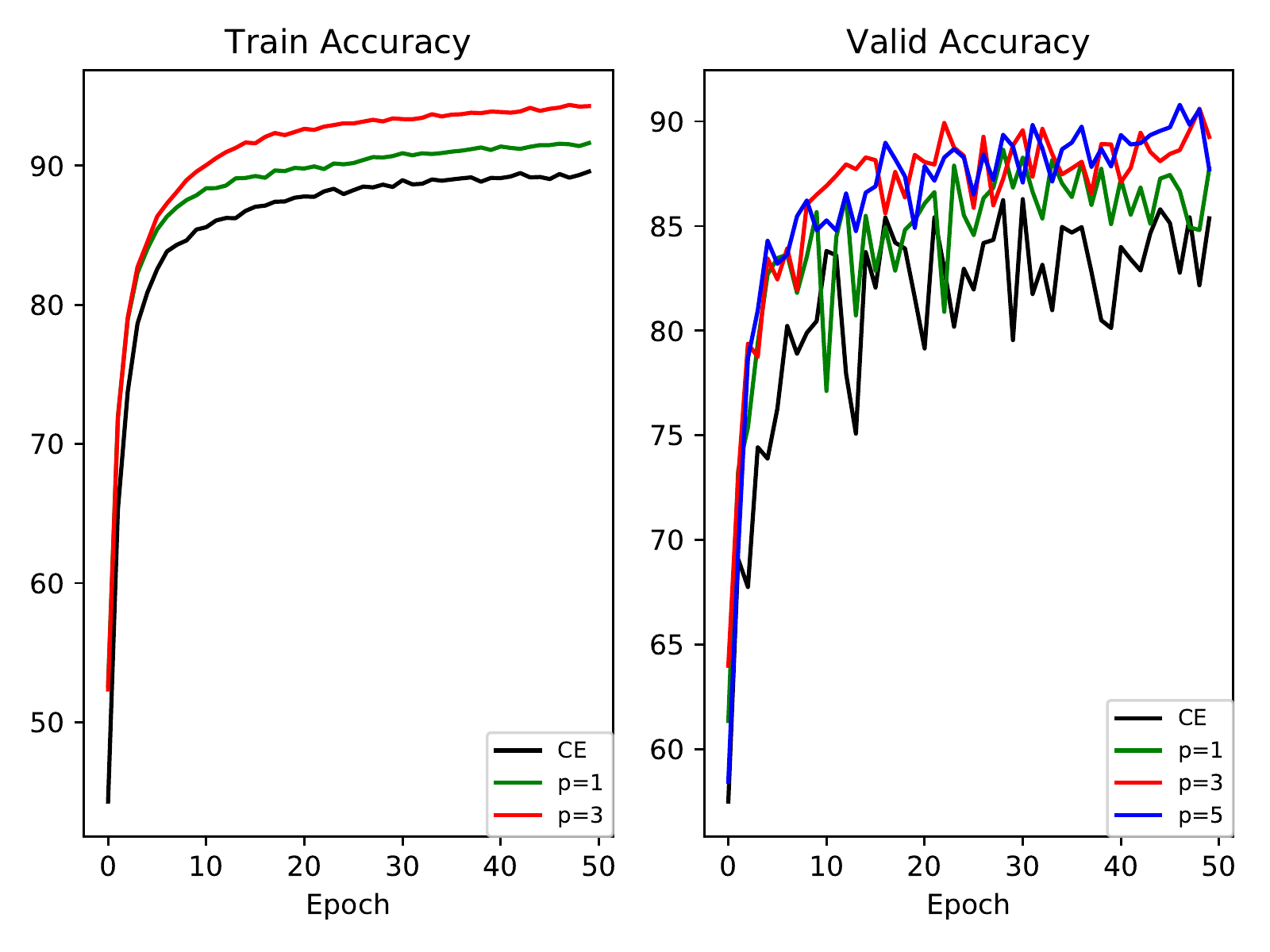} 
			\includegraphics[width=0.49\linewidth]{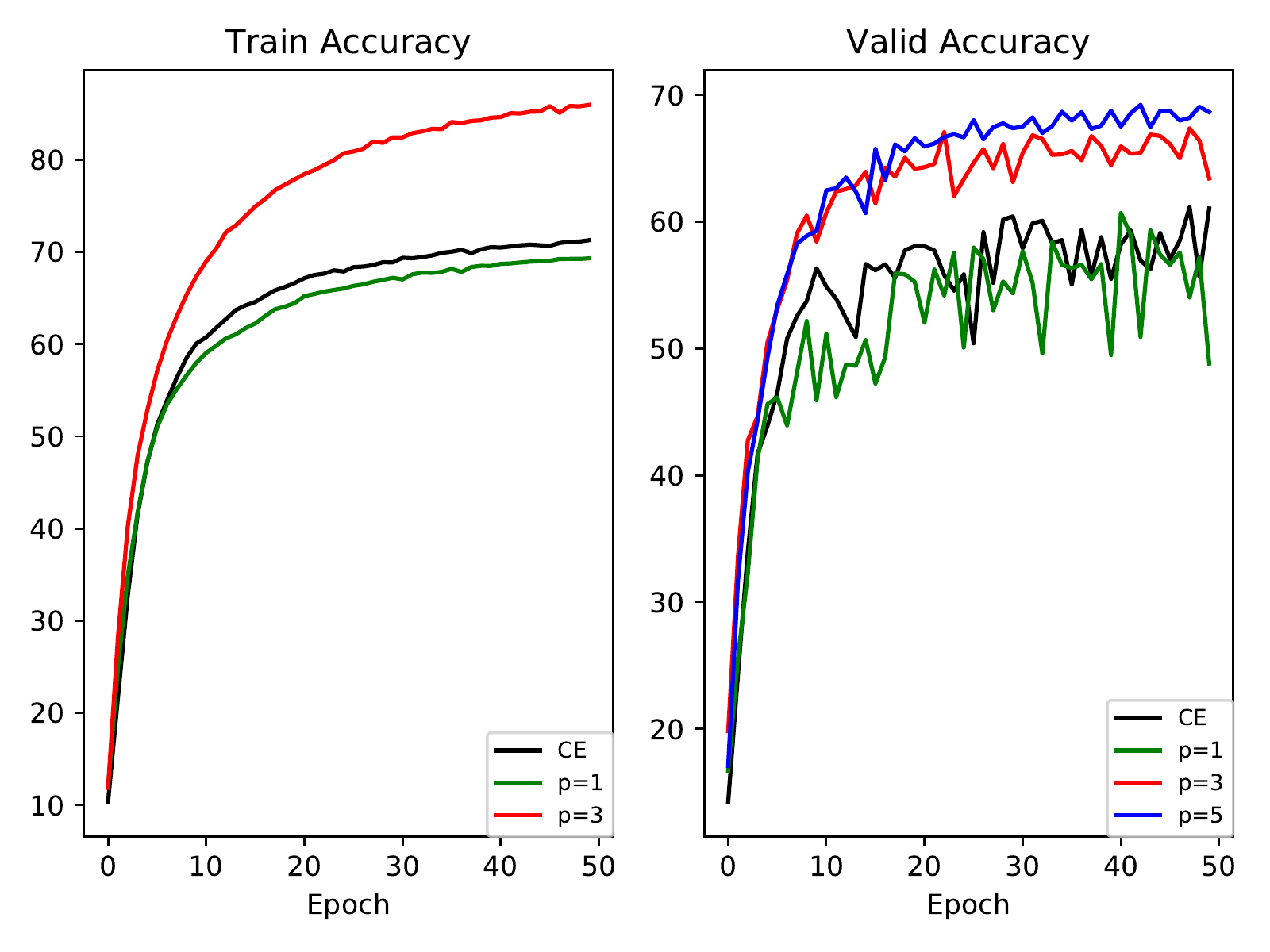} 
		\caption{ \textbf{(Top)} Residual Networks on \textbf{(Left)} CIFAR-10 and \textbf{(Right)} CIFAR-100. Regularized parameter in CE and linear method (i.e., $p=1$) is $5 \times 10^{-4}$. \textbf{(Bottom)} Densenet on \textbf{(Left)} CIFAR-10 and \textbf{(Right)} CIFAR-100. \textbf{(Left)} In single-loss and nonlinear collaborative schemes, regularization parameter is $1 \times 10^{-4}$, while in linear scheme (i.e., $p=1$), it is $2 \times 10^{-4}$.  \textbf{(Right)} In nonlinear collaborative scheme, regularization parameter is $5 \times 10^{-5}$, while in single-loss and linear schemes, it is $5 \times 10^{-4}$. $\epsilon=1 \times 10^{-4}$. }
		\label{fig:exp}
\end{figure}

General experiments, which are with original labels, on Residual Networks and Densenet are shown in Figure \ref{fig:exp}. From the train and validation accuracy, we can see that the linear scheme only compare favorably, even inferior, to the typical method with CE, both in Residual Networks and Densenet. However, our nonlinear collaborative scheme leads to best performance. Moreover, owing to the less oscillations, we argue that the nonlinear scheme is more robust. 


It should be mentioning that, in our experiments:  

\begin{itemize}
\item Owing to previous experimental results on MSE for classification \cite{golik2013cross,janocha2017loss}, we utilize CE to initialize the whole training process in both linear and nonlinear-collaborative scheme. After the training becomes stable, the objective function will be changed as the combination between MSE and CE \footnote{Similar training process can be found in \cite{golik2013cross,janocha2017loss,xu2016multi}. Effectiveness of this method is also interpreted in Appendix C intuitively. }. 

\item In multi-loss and nonlinear collaborative schemes, we need to add noise $ \mathcal{N}(0,\epsilon^2)$ in gradient to avoid overfitting.

\item Choise of optimizer does not affect the comparing results.  

\item We do not compare the values of different objective functions, which is an obvious result owing to the comparison between formulations and it does not make sense for showing actual training results. 

\item Based on the experiments in Refs. \cite{mse2012glcm,mse2013supervised,mse2015diagnosis,mse2016domain}, we do not worry the classification results when the objective function refers to MSE. In practice, we believe that the experimental results will be changed with different loss functions chosen.  
\end{itemize}

\subsection{Random Labels:} 

\begin{figure}[!htbp]
		     \centering
		\includegraphics[width=0.49\linewidth]{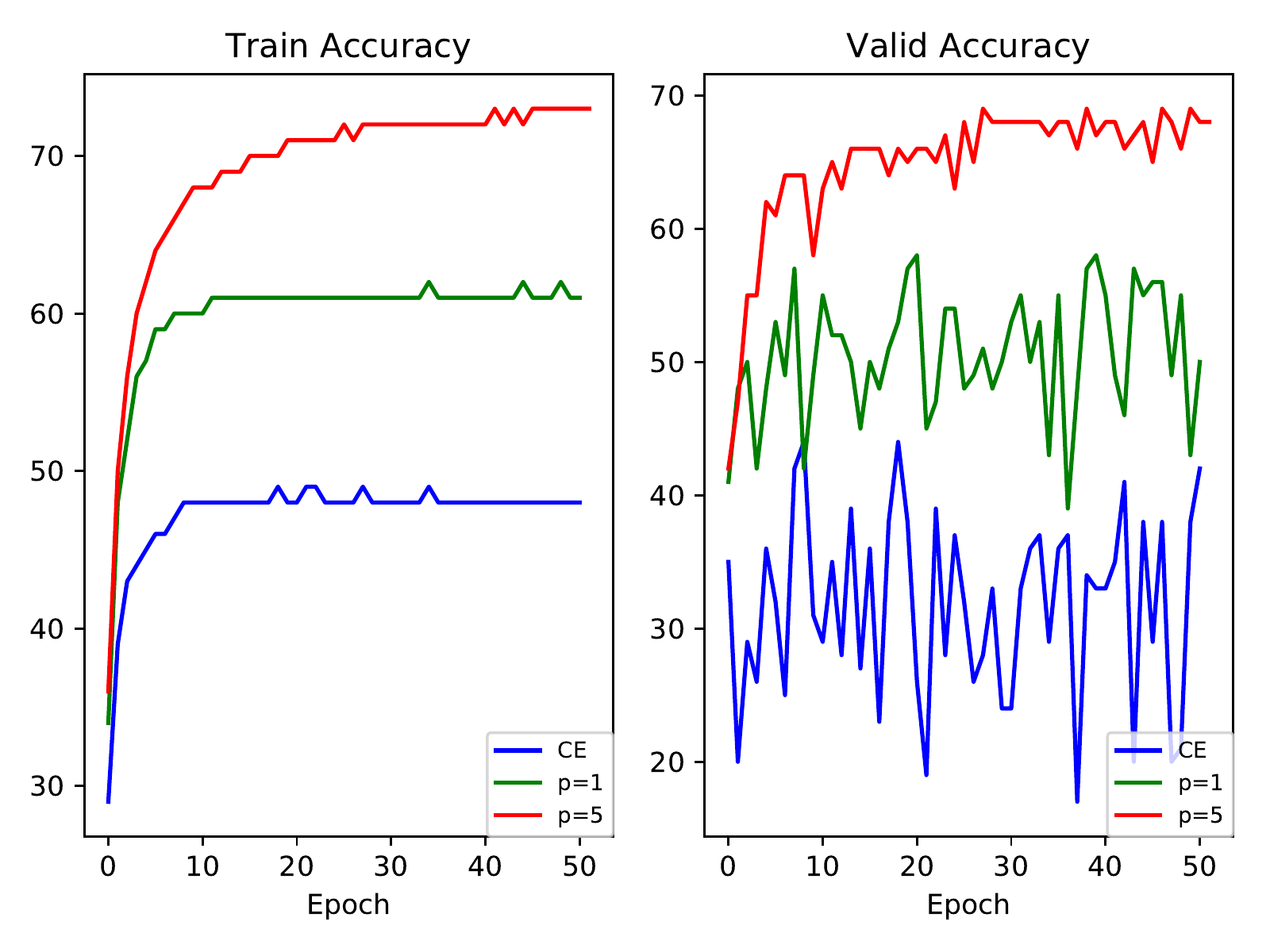} 
			\includegraphics[width=0.49\linewidth]{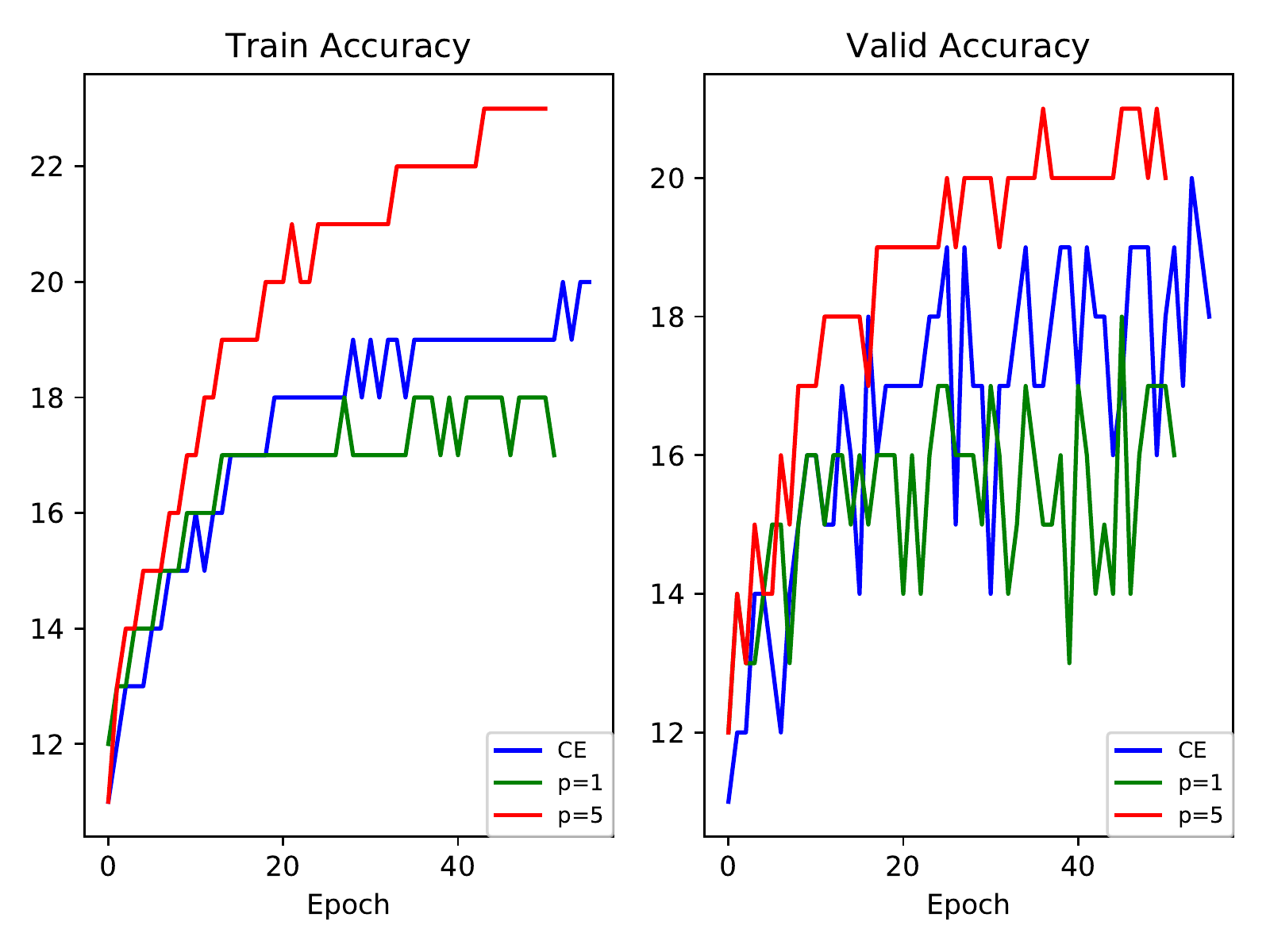}
            \includegraphics[width=0.49\linewidth]{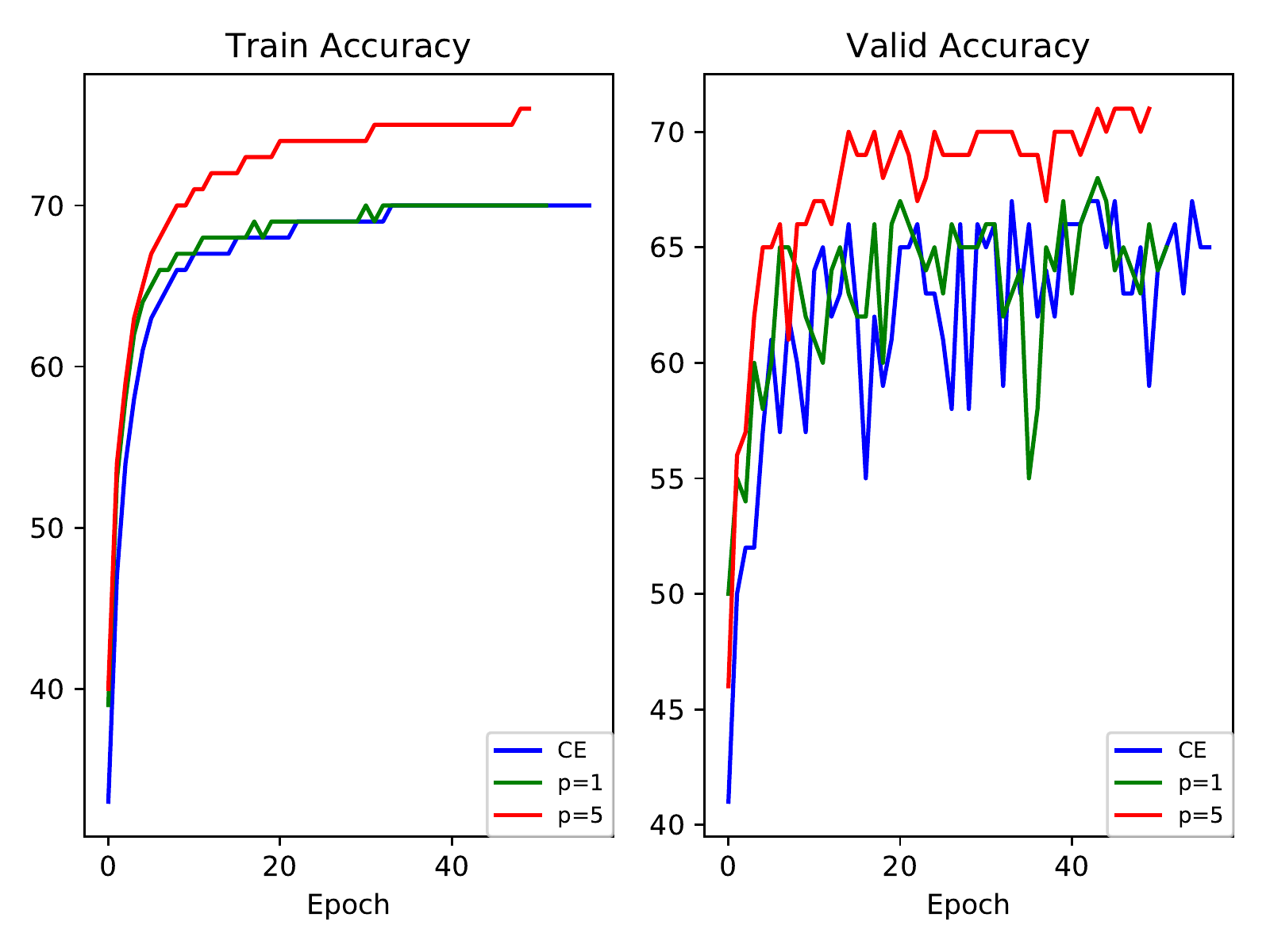} 
			\includegraphics[width=0.49\linewidth]{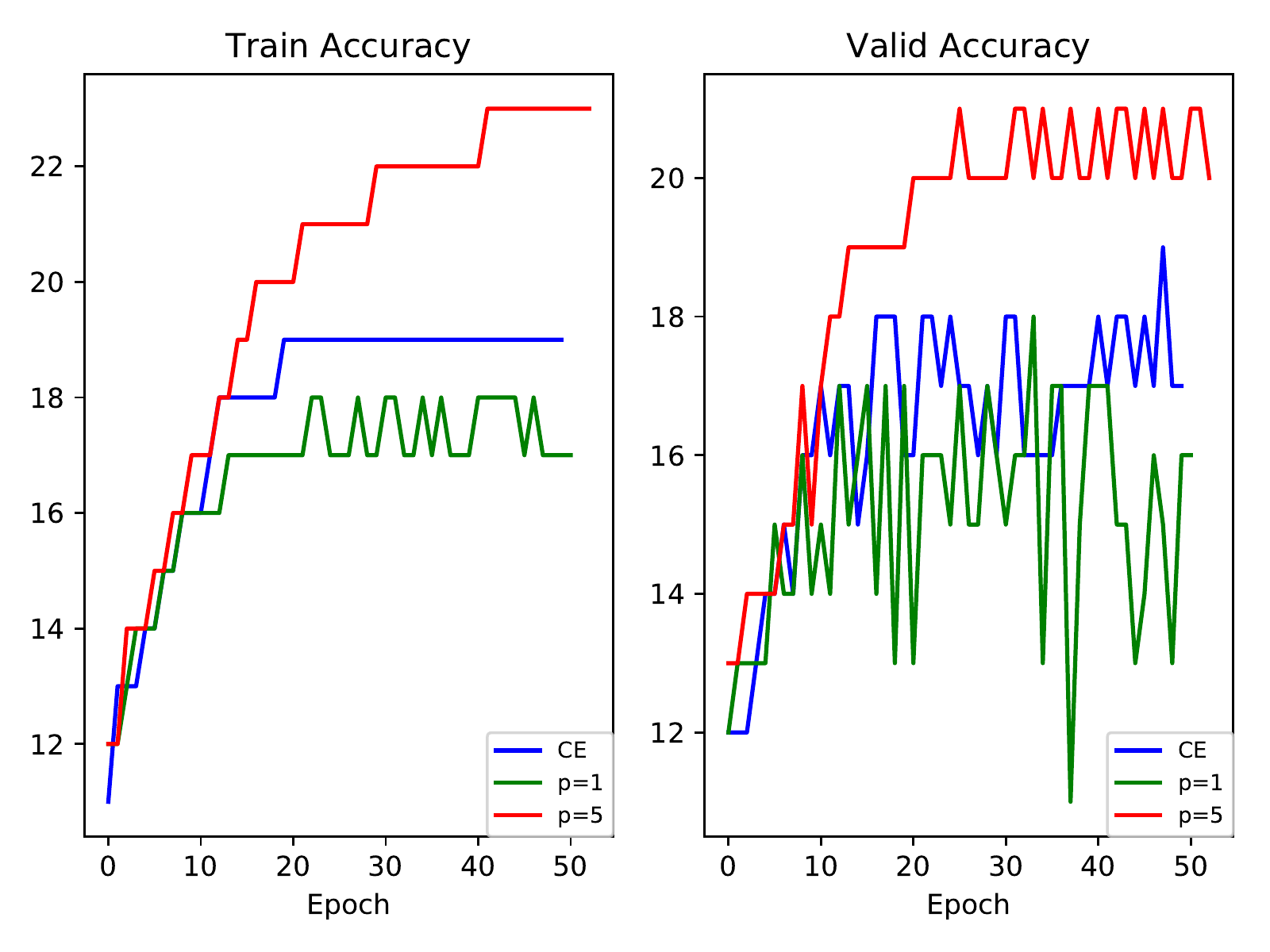} 
		\caption{ \textbf{(Top)} Residual Networks on CIFAR-10. \textbf{(Left)} Level of randomization is $20\%$, and regularized parameters in single-loss, multi-loss and nonlinear collaborative schemes, respectively, are $0.005$, $0.0001$ and $0.0002$. \textbf{(Right)} Level of randomization is $80\%$, and regularized parameters in single-loss, multi-loss and nonlinear collaborative schemes, respectively, are $0.0005$, $0.0001$ and $0.0001$. 
        \textbf{(Bottom)} Densenet on CIFAR-10 with  ifferent levels of random labels: \textbf{(Left)} Level of randomization is $20\%$, and the regularized parameters in single-loss, multi-loss and nonlinear collaborative schemes, respectively, are $0.0005$, $0.0003$ and $0.0001$. \textbf{(Right)} Level of randomization is $80\%$, and regularized parameters in single-loss, multi-loss and nonlinear collaborative schemes, respectively, are $0.0005$, $0.0003$ and $0.0001$.  }
		\label{fig:exp_random}
\end{figure}

Inspired by \cite{understanding2016}, to gain further insight into the comparison between our proposed scheme and previous ones, we experiment with different levels of randomization for labels on CIFAR-10 to explore the generalization ability of our nonlinear collaborative scheme, as shown in Figure \ref{fig:exp_random}.


Obviously, it is found that although the train and validation accuracy are both decreasing owing to the randomization of labels, the nonlinear collaborative scheme still performs superior to the multi-loss (i.e., linear) and single-loss schemes, with larger accuracy and relatively smooth training curves which denotes, respectively, better generalization ability and stronger robustness.   

\section{Conclusions}\label{sec:con}

We have proposed a \textit{nonlinear collaborative scheme} for network training. Its key technique lies in the construction of \textit{ a nonlinear collaborative objective function} using different typical loss functions, such as MSE, CE and JSD, to name a few. It is a completely different research line in comparison with the previous ``single-loss scheme" and ``linear scheme".

Its advantages have been demonstrated from two points, either of which corresponds to one step of relaxation during network training:
\begin{itemize}
\item On account of the relaxation from empirical zero $L_{0}$ risk to such computational surrogates like MSE and CE, we have demonstrated that our nonlinear collaborative objective function can serve as a better surrogate since it can balance between more choices for gradient and searching cost, resulting in a more adaptive training process. 

\item  In terms of the relaxation from text accuracy to train accuracy, we have proved that minimizing the nonlinear collaborative objective function is equivalent to minimizing a PAC-Bayes prior, in the context of a differentially private prior.  
\end{itemize}
Furthermore, we have proved that our proposed algorithm lead to a tighter PAC-Bayes bound, comparing with the entropy-SGD algorithm \cite{entropy_sgd,sgd_pac2017entropy}.
Our experimental results in Residual Networks and Densenet on CIFAR-10 and CIFAR-100 have shown that our nonlinear collaborative algorithm performs superior to both single-loss and linear scheme, owing to the train and validation accuracy and robustness. It should be noted that this adaptive nonlinear collaborative scheme is related to the multi-objective optimization based on decomposition \cite{zhang2007moea,zhang2009performance,zhang2010expensive}, thus several improvements can be considered in future work.

\section*{Acknowledgment}

The authors would like to thank all the members in research groups. 

\section*{Appendix}
\subsection{Method for Uniform Dimension}

To avoid the effect of different dimensions of $L_{mse}(x)$ and $L_{ce}(x)$, we give a method to uniform them. Firstly, we given the respective Gibbs distributions: 
\[ P_{mse} = \mathcal{Z}_{mse}^{-1} \mathrm{e}^{-L_{mse}(x)},\ \ \ 
P_{ce} = \mathcal{Z}_{ce}^{-1} \mathrm{e}^{-L_{ce}}, \]
with $\mathcal{Z}_{mse}$ and $\mathcal{Z}_{ce}$ as normalization parameter (partition function). 

Thus, the loss function from the view of Gibbs distribution can be given as 
\begin{align*}
& \tilde{L}_{mse} = \log P_{mse} = \log \mathrm{e}^{-L_{mse}(x)} + \log \mathcal{Z}_{mse}^{-1} = - L_{mse}(x)+C_1, \\
& \tilde{L}_{ce} = \log P_{ce} = \log \mathrm{e}^{-L_{ce}} + \log \mathcal{Z}_{ce}^{-1} = -L_{ce}+C_2,  
\end{align*}
with $C_1$ and $C_2$ as two integral constants. Clearly, dimensions of $\tilde{L}_{mse}$ and $\tilde{L}_{ce} $ are consistent. Then, we can use $\tilde{L}_{mse}$ and $\tilde{L}_{ce} $ to replace $L_{mse}$ and $L_{ce}$, respectively. Thus, no matter whether different loss functions have the same d

\subsection{Differential Privacy}

Here, to investigate that we do not need to worry whether the prior of PAC-Bayes bound is dependent on samples/observations or not, we introduce the definition of differential privacy and a standard approach.   

\textbf{$\epsilon$-differentially private \cite{dwork2008differential,sgd_pac2017entropy}}: Let $\epsilon$  be a positive real number, $\mathcal{A}$ be a randomized algorithm that takes a dataset as input (representing the actions of the trusted party holding the data), and $im \mathcal{A}$ be an image of $\mathcal{A}$. $\mathcal{A}$ is $\epsilon$-differentially private if for all datasets $D_1$ and $D_2$ that differ on a single element (i.e., the data of one person), and all subsets $S \in im \mathcal{A}$, we have $\mathbb{P}(\mathcal{A}(D_1)\in S) \leq e^{\epsilon} \mathbb{P}(\mathcal{A}(D_2)\in S)$. 

\textbf{Lemma 1 \cite{dwork2008differential,sgd_pac2017entropy}:} The standard approach for optimizing a data-dependent objective in a private way is to use the exponential mechanism. 

In the context of optimizing entropy $S$ in (10), the exponential mechanism corresponds to optimize $P_{\mathrm{exp}(S)}$. That is to say, $P_{\mathrm{exp}(S)}$ servers as the differentially private data-dependent prior \cite{dwork2008differential,sgd_pac2017entropy}.

\ifCLASSOPTIONcaptionsoff
  \newpage
\fi

\medskip

\small

\bibliographystyle{ieeetr}
\normalem
\bibliography{sample}

\end{document}